\documentclass[10pt,twocolumn,letterpaper]{article}

\usepackage{wacv}                         

\usepackage{xcolor}
\definecolor{wacvblue}{rgb}{0.21,0.49,0.74}
\usepackage[pagebackref,breaklinks,colorlinks,allcolors=wacvblue]{hyperref}

\usepackage{graphicx}
\usepackage{amsmath}
\usepackage{amssymb}
\usepackage{booktabs}
\usepackage{wrapfig}
\usepackage{pifont}
\newcommand{\cmark}{\ding{51}}%
\newcommand{\xmark}{\ding{55}}%
\usepackage{tabularx}
\usepackage{array}
\usepackage{float}
\usepackage{booktabs, multirow}
\usepackage{bigstrut}
\usepackage{comment}
\usepackage{arydshln}

\def\wacvPaperID{92} 
\def\confName{WACV}
\def\confYear{2026}

\title{Hierarchical Instance Tracking to Balance Privacy Preservation\\ with Accessible Information}

\author{
\begin{tabular}{c}
Neelima Prasad\textsuperscript{1},
Jarek Reynolds\textsuperscript{1},
Neel Karsanbhai\textsuperscript{1},\\
Tanusree Sharma\textsuperscript{2},
Lotus Zhang\textsuperscript{3},
Abigale Stangl\textsuperscript{4},\\
Yang Wang\textsuperscript{5},
Leah Findlater\textsuperscript{3},
Danna Gurari\textsuperscript{1} \\
\small
[1] University of Colorado Boulder \quad
[2] Pennsylvania State University \quad
[3] University of Washington \\
\small
[4] Georgia Institute of Technology \quad
[5] University of Illinois at Urbana-Champaign
\end{tabular}
}

\begin{document}
\maketitle

\begin{abstract}
We propose a novel task, hierarchical instance tracking, which entails tracking all instances of predefined categories of objects and parts, while maintaining their hierarchical relationships.  We introduce the first benchmark dataset supporting this task, consisting of 2,765 unique entities that are tracked in 552 videos and belong to 40 categories (across objects and parts). Evaluation of seven variants of four models tailored to our novel task reveals the new dataset is challenging. Our dataset is available at \href{https://vizwiz.org/tasks-and-datasets/hierarchical-instance-tracking/}{this URL}.
\end{abstract}
\section{Introduction}
\label{sec:introduction}
Many people use camera-based services to stream videos showing their daily activities.  For example, blind individuals regularly use them to learn about their visual surroundings~\cite{tseng2024biv, zhang2024designing, sharma2023disability, samson2024privacy}, including with Be My Eyes, Aira, and Envision AI.  A growing number of sighted users are also using extended reality devices to enrich their daily viewing experiences, including with Meta's Orion glasses, VITURE's XR glasses, and Apple's Vision Pro.  A key challenge for such video-based services is how to balance preserving privacy with retaining useful data.  

Two types of scenarios underscore the tension between privacy preservation and data retention.  \emph{First}, is when a person shares their video feed with another person.  This is common for blind people, who for example may want human confirmation regarding the required dosage for their prescribed medication in a particular pill bottle without revealing their name and address.  AI could assist by either (1) obfuscating everything except the part of interest (e.g., dosage information) or (2) obfuscating only private categories (e.g., name and address).  The \emph{second} scenario is when a person shares video with a service provider that subsequently saves the data.  It is common for blind people to share their private information with companies as a lesser evil to not learning about their visual surroundings~\cite{gurari2019vizwiz}, and in such cases obfuscating only the private categories (e.g., name and address) would preserve users' privacy while maintaining much of the utility of saved data for downstream purposes (e.g., training AI models).  Importantly, for both these scenarios, an incorrect segmentation in even a single video frame would mean that information a user wants to conceal is revealed.

Addressing the need to balance preserving privacy with retaining useful data, we propose a novel task we call \emph{hierarchical instance tracking}.  It entails identifying and tracking all instances of predefined categories of \emph{objects} and their \emph{parts}, while maintaining their \emph{hierarchical} relationships.  This task unifies two problems historically examined independently: \emph{video instance segmentation} (i.e.,tracking in videos all instances of predefined categories of \emph{objects}) and \emph{part segmentation} (i.e., locating in images all instances of predefined categories of \emph{parts of objects}).

\begin{figure*}[ht!]
    \centering
    \includegraphics[width=0.9\textwidth]{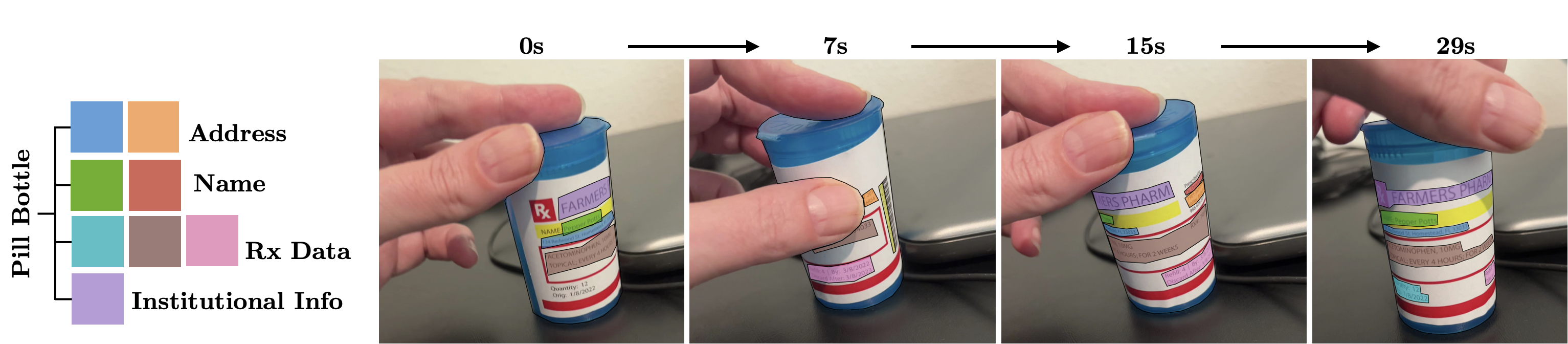}
    \caption{Example from our BIV-Priv-HIT dataset showing ground truth annotations we collected to track a pill bottle and its private parts. The legend on the left indicates the semantic labels and instance colors used to overlay tracked entities in the video frames.}
    \label{fig:motivation}
\end{figure*}

We introduce the first publicly-available dataset supporting our novel task, which includes annotations for tracking semantically-labeled objects and their parts using masklets (i.e., tracked segmentation masks).  Notably, this is also the first publicly-available dataset to even semantically track just parts alone.  We create the dataset by annotating 552 publicly-available videos taken by people with vision impairments of private content, which is called BIV-Priv~\cite{sharma2023disability}.  For each video, we segmented and tracked every object and its parts that belong to 40 semantic categories, resulting in tracks for 2,765 entities with 537 objects and 2,228 parts. We call the resulting annotated dataset \textbf{BIV-Priv-HIT}, reflecting the task of \textbf{H}ierarchical \textbf{I}nstance \textbf{T}racking (\textbf{HIT}).


Next, we analyze how BIV-Priv-HIT compares to eight existing datasets for entity tracking and hierarchical segmentation.  We show existing datasets \emph{cannot support our novel task} because (1) none track parts \emph{with} semantic labels and (2) none simultaneously track an object and its embedded parts, permitting the same pixel to belong to multiple semantic categories. We also show the new dataset \emph{fills important gaps of existing datasets} and so supports developing more generalized algorithms. For instance, videos in BIV-Priv-HIT are orders of magnitude longer, ranging from approximately double to 11 times as long as videos in existing datasets.  Additionally, segmentations in our dataset are unlike those in existing datasets including because (1) parts tend to contain text, (2) parts tend to have boundaries that are smoother and more elongated, at times even resembling a line segment,  and (3) objects occupy much larger portions of video frames.  This is exemplified in \textbf{Figure~\ref{fig:motivation}}.
 
Finally, we evaluate seven variants of four top-performing models for three related tasks---hierarchical image segmentation, video object segmentation, and video instance segmentation---on our dataset, after repurposing them for the task of hierarchical instance tracking. This effort includes introducing a new evaluation metric tailored to our tracking task. We found that all models perform poorly overall, especially for tracking parts but more generally for locating all small entities.  In addition, the models are inefficient, as they require ad hoc workarounds involving multiple inference passes to perform our task. These findings underscore the value of our new dataset in supporting the AI community to tackle a new challenging problem.  

We expect our dataset challenge will inspire new algorithmic designs for handling a greater diversity of real-world challenges within a single model.  Success in this work could benefit other privacy-preserving applications, such as for individuals conversing with video streaming services (e.g., Zoom, WhatsApp), and robots navigating environments (e.g., emergency responders sent to crumbling/burning buildings).  Success can also benefit other video-based applications by infusing finer-grained part-level understanding, including for robotics manipulation tasks, video editing, video retrieval, and pose estimation.
\section{Related Work}
\label{sec:related}

\paragraph{Part and Hierarchical Segmentation Datasets.} 
The recent successes of segmentation models at locating \emph{objects} belonging to pre-defined categories has prompted a shift of focus for the community to instead segment more challenging \emph{part-level categories}. This shift, which began in the mainstream computer vision community around 2017, has been inspired and enabled by new datasets that provide segmentations showing how objects are hierarchically decomposed into their nested parts~\cite{zhou2017scene, liang2018look, zhao2018understanding,gong2018instance, wah2011caltech,zheng2018modanet,song2019apollocar3d, jia2020fashionpedia,de2021part,ramanathan2023paco, wei2024ov,chen2014detect,he2022partimagenet}.  Our work complements this literature by providing the first dataset containing semantic part segmentations as well as hierarchical segmentations for \emph{tracking content in videos}.  Additionally, parts in our dataset fill a gap of existing part-based datasets by exhibiting unique characteristics, including their tendency to contain text and so have smoother, more elongated boundaries.

\vspace{-1em}\paragraph{Tracking Datasets.} 
Two popular types of entity tracking datasets exist.  \emph{Video object segmentation} (VOS) datasets provide masks of entities tracked across all video frames (e.g., SA-V~\cite{ravi2024sam}, DAVIS \cite{pont20172017}, YouTube-VOS~\cite{vos2022}), while \emph{video instance segmentation} (VIS) datasets also require labeling the category for each tracked entity (e.g., YouTube-VIS~\cite{Yang2019vis}).  Most similar to our work is the SA-V~\cite{ravi2024sam} VOS dataset because it is the only other dataset that contains masklets (i.e., tracked masks) for \emph{parts}.  However, the SA-V dataset does not (1) provide semantic labels or (2) specify whether a masklet is for an object or a part (inferring this is non-trivial). Our work fills both these gaps.


\vspace{-1em}\paragraph{Models for Segmentation Tracking and Hierarchical Segmentation.}
None of the models for tracking or hierarchical segmentation support our proposed task.  For instance, video instance segmentation (VIS) models can be trained and applied for our target semantic object and part categories, but they cannot achieve this in a single inference pass since they do not permit the same pixel to belong to multiple semantic categories.\footnote{An orthogonal line of research focuses on part-based tracking.  While these methods also track ``parts", their definition differs: in these approaches, parts are treated as  appearance cues on objects (i.e., patches) that enhance the robustness of object tracking, rather than as semantic part categories with distinct identities~\cite{yao2013part,lukežič2016,akin2016,Yao2017,huang2017visual,cheng2018,de2018part,burceanu2018learning,zhang2023multiple}.}  Similarly, video object segmentation (VOS) models~\cite{ravi2024sam} don't permit the same pixel to belong to multiple semantic categories and so would require multiple inference passes to support our task.  Extending beyond VIS' limitations, VOS models also ignore semantics and require human annotation at the first appearance of each entity to track them.  An alternative approach could be to apply hierarchical instance segmentation models~\cite{wang2024hierarchical} to every video frame to locate objects and parts, however such models ignore the fundamental concept of preserving ``identities" over time and so would necessitate extra complexity to associate segmentation masks across video frames. Despite modern models' limitations, we benchmark them using ad hoc workarounds to highlight their potential value as a foundation for addressing our novel task.  While experimental results reveal all types struggle, underscoring our dataset offers a challenging problem for the research community, VOS models offer the greatest promise.


\vspace{-0.75em}\paragraph{Datasets Originating from Blind Individuals.} 
This work also contributes to the movement in creating benchmark datasets where visual content originates from blind individuals.  Most work focuses on images~\cite{gurari2018,kim2019korean,gurari2019vizwiz,bhattacharya2019does,gurari2020,chiu2020assessing,zeng2020vision,chen2022grounding,tseng2022vizwiz,bafghi2023new,chen2023vqa,tseng2024biv,reynolds2024salient,huh2024long,chen2025acknowledging}, with VizWiz~\cite{gurari2018} pioneering this direction in 2018, yet none provide hierarchical segmentations.  Other efforts focus on videos~\cite{orbit,EgoBlind, BLVnavigation,li2025exploring}, yet none provide masklet annotations.  Our work fills both gaps, contributing to the broader goal of designing more inclusive AI models that address the interests of blind people.


\section{Hierarchical Instance Tracking Dataset}
\label{sec:dataset}
We now introduce BIV-Priv-HIT, the first dataset that supports hierarchical instance tracking (and part tracking).

\subsection{Dataset Creation}

\paragraph{Video Source.}
We leverage the 552 publicly-available videos from BIV-Priv~\cite{sharma2023disability}, which were captured by 26 blind photographers of 16 private object categories~\cite{tseng2024biv} (e.g., credit cards). Importantly, none of the private content was pertinent to the photographer and instead originated from the datasets' authors with Institutional Review Board approval. Each photographer was instructed to capture an approximately 25 second clip for each type of private object twice, once with it positioned in the background to mimic \emph{accidental} privacy disclosures and once in the foreground to mimic \emph{intentional} privacy disclosures, emulating what was previously observed in authentic use cases~\cite{gurari2019vizwiz}.  

\vspace{-1em}\paragraph{Hierarchical Category Selection.} 
We identified 24 \emph{part} categories to associate with the 16 \emph{object} categories established when curating the BIV-Priv videos~\cite{tseng2024biv}.  We developed the part taxonomy to capture three tiers of common privacy concerns~\cite{li2020towards, PII_Glossary, pei2023tale, xu2024dipa2}:
\begin{itemize}
    \item \emph{Personally Identifiable Information} (PII): information directly revealing an individual's identity, such as names, account numbers, and credit card numbers. 
    \item \emph{Quasi-personally Identifiable Information}: information that can indirectly reveal an individual's identity, such as addresses and job titles. 
    \item \emph{Sensitive Information}: not tied to a person's identity, but information a person might not want shared with others. 
\end{itemize}

\noindent    
A list of all object categories and their associated privacy categories is shown in \textbf{Figure~\ref{fig:box-plot-amount-of-each-part}}.  

\begin{table*}[t!]
    \centering
    \scriptsize
    \begin{tabular}{lcccccc}
    \hline
        & Ours & ADE20K \cite{zhou2019semantic} & PACO-EGO4D \cite{ramanathan2023paco} & PACO-LVIS \cite{ramanathan2023paco} & PartImageNet \cite{he2022partimagenet} & PASCAL-Part \cite{chen2014detect} \\ \hline
        Tracking & {\color{green}\cmark} & {\color{red}\xmark} & {\color{red}\xmark} & {\color{red}\xmark} & {\color{red}\xmark} & {\color{red}\xmark} \\ 
        \# of Images & 11.1K & 27.6K & 26.3K & 57.6K & 24K & 19.7K  \\ 
        \% of Images with Object Masks & 91\% & 100\% & 90.9\% & 91.5\% & 100\% & 96.4\%  \\ 
        \% of Images with Part Masks & 67.2\% & 45.7\% & 90.8\% & 91.5\% & 100\% & 96.4\%  \\ 
        \% of Objects with Part Masks & 73.8\% & 13.6\% & 87.2\% & 76.4\% & 100\% & 80\%  \\ \hline
    \end{tabular}
    \vspace{-0.5em}
    \caption{Comparison of the composition of our dataset to existing hierarchical object-part segmentation datasets. While existing datasets hierarchically segment objects in images, BIV-Priv-HIT is the first to hierarchically segments objects across videos' frames.}
    \label{table:part-image-datasets}    
\end{table*}

\vspace{-1em}\paragraph{Annotation Collection.} 
For every video, we tracked every instance of each object and part category in our taxonomy using masklets.  This involved a three-step process.  

First, we constrained the annotation of masklets to only the clips of videos with target objects visible.  To achieve this, three in-house annotators contributed by indicating the start and end frames where any target object category appeared (along with the detected category).  For quality control, we had each video annotated by two people and then annotation differences were resolved via group discussions.  

Next, we collected segmentations for every instance of the target objects and parts.  We hired 25 trusted crowdworkers\footnote{We vetted the crowdworkers through their involvement in creating for us $\sim$40,000 object and part segmentations for other datasets.} from Amazon Mechanical Turk (AMT) to complete this using a home-grown interface which presents each frame and then has the annotator sequentially segment a specified object category followed by every visible instance of specified part categories. To accommodate occlusions fragmenting entities into multiple parts, we included a feature that enables creating multiple polygons when segmenting a single entity.  We supported high annotation quality via on-boarding `warm up' tasks, detailed instructions, live `office hours' during annotation deployment periods for answering questions, and phased task rollouts to enable time for continuous inspection of submitted annotation results and worker feedback.  Additionally, for each video frame, we collected annotations from two crowdworkers and then used their similarity to establish high-quality ground truth instance segmentations (as described in the supplementary materials).  We chose to annotate every 40th video frame to balance annotating enough frames to be comparable in size to existing VOS datasets (see \textbf{Figure~\ref{fig:box-plot-amount-of-each-part}}) while skipping  neighboring frames with very similar appearances.  In total, 11,165 frames were annotated.  This culminated in 10,165 object segmentation masks and 22,037 part segmentation masks, with 8.9\% (i.e., 1,000) of video frames lacking segmentation masks.  Cumulatively, the crowdworkers took 1,820 annotation hours (i.e., 45.5 40-hour work weeks) to complete the annotations.  

Finally, in-house annotators associated the annotation masks belonging to the same entity across video frames and assigned unique instance IDs to each resulting masklet (for objects and parts). This was achieved using a home-grown tool, with quality ensured by having all resulting masklets verified by a second author. Cumulatively, these assocation annotations were completed in approximately 150 hours, which translates to an average of just over 3 minutes to associate all masks for each of the 2,765 tracked entities.


\vspace{-1em}\paragraph{Dataset Splits.}
We divided the videos into training, validation, and testing splits using a 60\% (327 videos), 15\% (87 videos), and 25\% (138 videos) split respectively. This resulted in the following number of annotated frames in the three splits: 6,690 in training, 1,680 in validation, and 2,795 in testing.  When creating the splits, we ensured that the blind videographers who initially recorded the videos did not appear across multiple datasets splits to prevent models from overfitting to features of specific photographers. 

\subsection{Dataset Analysis}
We now characterize BIV-Priv-HIT and how it compares to existing datasets.

\vspace{-1em}\paragraph{Baseline Datasets for Comparison.}
We chose to compare our dataset to existing datasets that support the two distinct problems our proposed task unifies into the same framework: hierarchical instance segmentation in images and entity tracking in videos.

For \emph{hierarchical segmentation} datasets, we chose those that similarly provide segmentations with semantic labels for objects and their nested parts.  We chose the following recent and popular datasets: PACO-LVIS~\cite{ramanathan2023paco}, PACO-EGO4D~\cite{ramanathan2023paco}, ADE20K~\cite{zhou2019semantic}, PartImageNet~\cite{he2022partimagenet}, and PASCAL-Part~\cite{chen2014detect}.

We chose \emph{entity tracking} datasets that similarly provide segmentation masks of entities throughout each video's frames to create \emph{masklets}.  This includes a popular video object segmentation dataset which provide masklets of objects \emph{without} associated semantic labels---DAVIS~\cite{pont20172017}---and a popular video instance segmentation dataset which provides masklets of objects \emph{with} associated semantic labels---YouTube-VIS~\cite{Yang2019vis, vis2022}.  Also included is the recent SA-V~\cite{ravi2024sam} dataset which is the first dataset to provide part-level masklets, although without semantic labels or explicit flags indicating whether a tracked entity is an ``object" or ``part".

\begin{table}[t!]
    \centering
    \scriptsize
    \begin{tabular}{l c c c c}
    \hline
     & Ours & DAVIS \cite{pont20172017} & YT-VIS \cite{vis2022} & SA-V \cite{ravi2024sam} \\ \hline
    Contains Semantics & {\color{green}\cmark} & {\color{red}\xmark} & {\color{green}\cmark} & {\color{red}\xmark} \\ 
    Part Tracking & {\color{green}{\color{green}\cmark}} & {\color{red}\xmark} & {\color{red}\xmark} & {\color{green}\cmark}* \\ 
    Hierarchy Tracking & {\color{green}\cmark} & {\color{red}\xmark} & {\color{red}\xmark} & {\color{red}\xmark} \\ 
    \# of Videos & 552 & 150 & 4,019 & 50.9K \\ 
    Mean Length (sec) & 27.9 & 2.4 & 5.31 & 14 \\ 
    \# Annotated Frames & 11,165 & 10,459 & 4,519 & 4.2M \\ 
    \# Instance Masks & 32,202 & 27.1K & 265.5K & 35.5M \\ 
    \# Unique Instances & 2,765 & 376 & 8,698 & 642.6K \\ 
    Disappearance Rate & 9\% & 16.1\% & 10.8\% & 42.5\% \\\hline
    \end{tabular}
    \vspace{-0.5em}
    \caption{Comparison of the composition of our dataset to existing entity tracking datasets. BIV-Priv-HIT is unique because it is the first to contain both semantic part annotations and hierarchical object-part instance segmentations as well as because it has longer video durations. (* flags that a dataset, in this case SA-V, lacks labels specifying whether any given mask is of an object or a part.)}
    \label{table:entity-tracking-datasets}
\end{table}
\vspace{-1em}\paragraph{Dataset Composition.}
We first characterize and compare BIV-Priv-HIT's overall composition to existing datasets.

\begin{figure*}[t!]
    \centering
    \includegraphics[width=\textwidth]{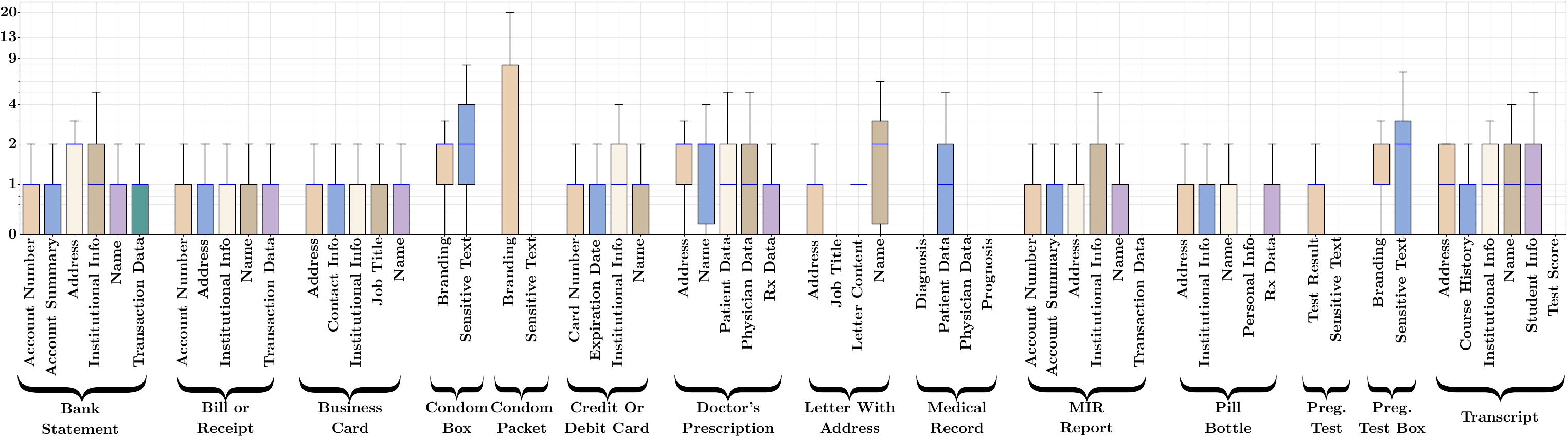}
    \vspace{-1.0em}
    \caption{Boxplot showing for the entire dataset how many of each type of part annotation is visible for each object category. Black centered lines represent medians, bottoms and tops of each box represent the 25th and 75th percentile values respectively, and the whiskers represent the most extreme data points not considered outliers. Two object categories are excluded due to zero or rarely annotated parts, resulting in insufficient data for a rendered box on the plot: ``Tattoo Sleeve" and ``Local Newspaper".}
     \label{fig:box-plot-amount-of-each-part}
\end{figure*}

\begin{figure*}[t!]
    \centering
    \includegraphics[width=\textwidth]{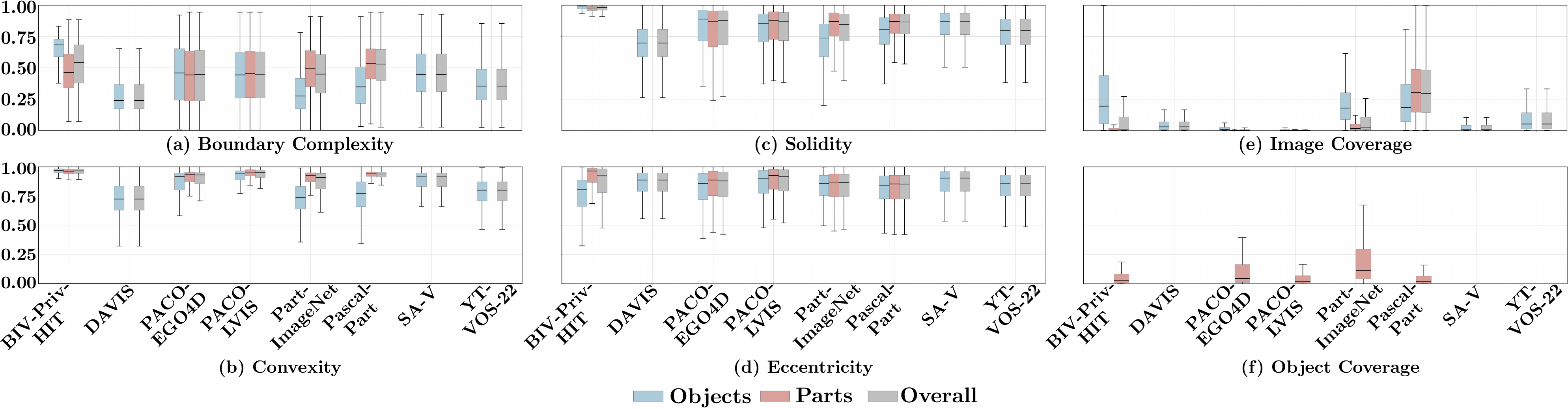}
    \vspace{-1.5em}
    \caption{Boxplots characterizing segmentations in our dataset and seven other datasets with respect to six metrics, overall as well as with respect to only objects and parts independently.  (Note: object coverage can only be computed for hierarchical segmentation datasets.)}
     \label{fig:spatial-bp}
\end{figure*}

We report in \textbf{Table~\ref{table:part-image-datasets}} a characterization of the \emph{hierarchical segmentations} for all relevant datasets.  As shown, BIV-Priv-HIT is unique because it is the only dataset to provide hierarchical segmentations for tracking entities in videos.  All other datasets only provide segmentations for images. BIV-Priv-HIT also tends to have a smaller prevalence of annotated images showing at least one object of interest (i.e., 91\%) and part of interest (67\%), an especially important feature since tracked entities can disappear by leaving the field of view or becoming occluded.

We report in \textbf{Table \ref{table:entity-tracking-datasets}} a characterization of all relevant \emph{entity tracking} datasets.  As shown, BIV-Priv-HIT is unique in three key ways. First, it is the only dataset to provide object and part segmentations with semantic labels.  Second, it is the first to track hierarchically decomposed objects (i.e., of objects and their nested parts).  Third, the typical video duration for BIV-Priv-HIT is orders of magnitude longer, with the mean video length of 27.9 seconds nearly double the mean length for SA-V~\cite{ravi2024sam} and over 11 times longer than the mean length for DAVIS~\cite{pont20172017}.  Beyond these unique features, BIV-Priv-HIT is largely comparable to existing datasets. For instance, it lies in the middle of the pack with respect to size (i.e., number of included videos, annotated frames, and instance masks) and the number of unique entity instances.  It also exhibits a similar prevalence of entity's disappearing in a video.

Next, we perform more fine-grained analysis of the prevalence of parts in BIV-Priv-HIT.  We observe that different object categories can contain different amounts of parts (boxplots provided in the Supplementary Materials), ranging from many (e.g., bank statements contain up to 6 parts) to few (e.g., local newspapers contain at most 1 part) to none (e.g., tattoo sleeves).  This underscores the value of our dataset in encouraging the design of models that can account for object decompositions of different complexities. Additionally, within a specific object category, we observe variability regarding how many parts and which types are visible (\textbf{Figure~\ref{fig:box-plot-amount-of-each-part}}). Consequently, our dataset will encourage models to rely on visual evidence in each frame rather than biases from typical object-part associations. 


\vspace{-1em}\paragraph{Segmentation Properties.} 
We also characterize the appearance of objects and parts and how they compare with segmented entities in related datasets\footnote{For datasets larger than BIV-Priv-HIT, we randomly sample 10,165 annotations for analysis to avoid excessive computational cost.  For datasets with part annotations, we sample objects with at least one part annotation. For datasets smaller than BIV-Priv-HIT we use the entire dataset.}.  To do so, we use the following metrics:

\begin{itemize}
    \item \textbf{Boundary Complexity}: ratio of a entity’s area to the length of its perimeter (\ie, isoperimetric quotient). Values range from 0 (highly jagged boundary) to 1 (circular).
    \item \textbf{Convexity}: ratio of the entity's convex hull's perimeter to the perimeter of the entity. Values range from 0 (significant concavities) to 1 (perfectly convex).
    \item \textbf{Eccentricity}: ratio of the distance between a segmentation's foci and the length of its central axis. Values range from 0 (circular) to 1 (line segment). 
    \item \textbf{Solidity}: ratio of the entity's area to the area of the entity's convex hull. Values range from 0 (fragmented or concave) to 1 (compact and solid).
    \item \textbf{Image Coverage}: ratio of the the image's pixels occupied by the entity (object or part). Values range from 0 (no image coverage) to 1 (complete image coverage).
    \item \textbf{Object Coverage}: ratio of the object's pixels occupied by all its nested parts. Values range from 0 (no object coverage) to 1 (complete object coverage).
\end{itemize}
\noindent

Results are shown in \textbf{Figure~\ref{fig:spatial-bp}}.  When comparing BIV-Priv-HIT to all the entity tracking and hierarchical segmentation datasets, we observe segmentation masks in our dataset are distinct for five of the six metrics (not object coverage).  We anticipate these distinctions, described below, will encourage the design of models that can generalize to a greater diversity of segmentation mask types.

With respect to the \emph{entity boundary}, we observe that the objects in BIV-Priv-HIT exhibit the least complexity (\textbf{Figure \ref{fig:spatial-bp}a}), convexity (\textbf{Figure \ref{fig:spatial-bp}b}), and solidity (\textbf{Figure \ref{fig:spatial-bp}d}).  This makes sense since our dataset focuses largely on human-made artifacts that typically have rectangular shapes, such as documents (i.e., pieces of paper), boxes, and pregnancy tests.  Such human-made artifacts, unless damaged, lack concavities and jagged edges.  

With respect to the \emph{entity shape}, objects tend to be more circular than elongated (e.g., line segment) while parts tend to be more elongated than circular compared to existing datasets (\textbf{Figure \ref{fig:spatial-bp}c}).  We attribute the latter observation to parts typically being textual information, which manifests as a line segment. 

With respect to \emph{entity size}, BIV-Priv-HIT's objects occupy the greatest diversity of sizes while parts occupy the least diversity of sizes compared to existing datasets (\textbf{Figure \ref{fig:spatial-bp}e}).  We attribute the former finding to the fact that the objects of interest were intentionally positioned both in the background and foreground of images by the photographers.  Moreover, objects can occupy larger portions of an image than observed in other datasets.  This finding aligns with prior work's findings~\cite{reynolds2024salient, tseng2022vizwiz, chen2022grounding}, which noted that people with vision impairments take close-up photographs of objects to better facilitate visual interpreters to recognize and so describe the visual content.

\vspace{-1em}\paragraph{Semantic Properties.} 
The 40 semantic categories in BIV-Priv-HIT share little overlap with all other datasets shown in \textbf{Tables \ref{table:part-image-datasets}} and \textbf{\ref{table:entity-tracking-datasets}}.\footnote{While SA-V could overlap with BIV-Priv-HIT due to its scale, SA-V excludes semantic labels preventing such comparison.}  We attribute this to existing datasets' focus on content lacking private information, in accordance with best practices for dataset creation to remove such content.  Our work, in contrast, \emph{centers} on private categories.  

Still, existing datasets can capture categories in our dataset with more abstract forms.  For example, `document' is a more general form of our `bank statement' category. Additionally, PACO-EGO4D, PACO-LVIS, and PartImageNet all feature a category label for `bottle' which shares a partial overlap with BIV-Priv-HIT's `pill bottle'; however, the other datasets focus at the part-level categories on the composition and anatomy of the bottle (e.g., cap, neck, body, and label) while  BIV-Priv-HIT focuses on the private contents of a pill bottle's label (e.g., address and prescription) \cite{ramanathan2023paco, he2022partimagenet}. PASCAL-Part also features a label for `card,' yet it makes no distinction as to what kind of card, such as credit card, playing card, business card, and so on \cite{chen2014detect}. Last, ADE20K shares the most partial overlap with BIV-Priv-HIT, with its category labels of bottle, card, bill, and document \cite{zhou2017scene}. However, these objects do not contain part-level data and again represent more abstract, non-privacy-centric forms of the objects. While the category alignment with our dataset is limited, we suspect the few similarities across categories could facilitate models trained on the more abstract categories to generalize to the more specific categories encountered in our dataset.
\section{Evaluating Hierarchical Instance Tracking }
\label{sec:metrics}
Given the novelty of our task, we introduce a metric that for assessing how well models can preserve the hierarchical structure between an objects and its parts throughout tracking.  Our key idea is to extend MOTA \cite{MOTA}, a standard metric for multi-object tracking: 

$$\text{MOTA} = 1-\frac{\sum_t FN_t+FP_t+IDSW_t}{\sum_t GT_t}$$

\noindent
where $t$ is the frame index, $FN$ are False Negatives, $FP$ are False Positives, $IDSW$ are identity switches, and $GT$ is the ground truth of one object. 



 




We call our new metric MOTA-H, to reflect that it calculates tracking object's \textbf{H}ierarchical compositions such that parts of an object should remain associated with that object over time.  Like MOTA, MOTA-H's score range from 1.0 for perfect tracking accuracy to negative infinity. Unlike MOTA, which determines detection matches between a prediction and ground truth using the overlap of bounding boxes, we instead use intersection over union (IoU) between segmentation masks, setting the IoU threshold to 0.5.  We then change how identity switches are calculated to incorporate hierarchical relations as follows:
\[
\textit{H-IDSW} = 
\begin{cases}
1, & \text{if predicted part ID changes } \\
1, & \text{if predicted parent object changes } \\
0, & \text{otherwise}
\end{cases}
\]

\noindent
resulting in the following equation for MOTA-H, where all scores are only measured for the parts:

$$\text{MOTA-H} = 1-\frac{\sum_t FN_t+FP_t+\textit{H-IDSW}_t}{\sum_t GT_t}$$

To also evaluate performance for tracking \emph{objects}, we include a MOTA variant we call MOTA-OBJ, where the only change is computing detection matches between predicted and ground truth \emph{objects} using IoU between segmentation masks (instead of the overlap between bounding boxes).
\section{Model Benchmarking}
\label{sec:benchmark}
We next benchmarked seven variants of four models.  All experiments were run on NVIDIA's Tesla A100 GPUs. 

\begin{table*}[t!]
    \centering
    \resizebox{\textwidth}{!}{
    \begin{tabular}{ l c c c c cc ccc ccc ccc }
        \toprule
        & \textbf{Cheats} & \textbf{Inf./Vid.} & \textbf{MOTA-H} & \textbf{MOTA-OBJ} & \multicolumn{3}{c}{{\bf J \& F}} & \multicolumn{3}{c}{{\bf AP}} & \multicolumn{3}{c}{{\bf AR}} \\ 
        \cmidrule(lr){6-8} \cmidrule(lr){9-11} \cmidrule(lr){12-14}
        \textbf{} & & \textbf{(Mean)} & & & \textbf{Total} & \textbf{Objects} & \textbf{Parts} & \textbf{Total} & \textbf{Objects} & \textbf{Parts} & \textbf{Total} & \textbf{Objects} & \textbf{Parts} \\
        \midrule
        HIPIE-R50 & {\color{red}\xmark} & 20.25 & -- & -- & 0.03 & 0.03 & 0.0 & 0.0 & 0.0 & 0.0 & 0.0 & 0.0 & 0.0 \\
        HIPIE-ViT & {\color{red}\xmark} & 20.25 & -- & -- & 0.03 & 0.03 & 0.0 & 0.0 & 0.0 & 0.0 & 0.0 & 0.0 & 0.0 \\
        \hdashline
        2 Mask2Formers         & {\color{red}\xmark}   &  2.00 &0.03  &0.12& 0.27&0.21&   0.07  & 0.41&0.25&0.11 &0.65&0.41&0.32 \\
        1 Mask2Former & {\color{red}\xmark} &1.00  & 0.00 &0.12   &0.21 &0.21&0.00         & 0.25 &0.25 &0.00          & 0.41 &0.41 &0.00 \\
        \hdashline
        XMem++ & {\color{green}\cmark} & 6.05  & 0.47&0.71& 0.73 &0.77&0.69         & 0.71  &0.74& 0.66       & 0.79&0.82&0.75 \\
        SAM-2 & {\color{green}\cmark} & 6.05 & 0.39 & 0.54 & 0.58 & 0.73 & 0.53 & 0.58 & 0.82 & 0.53 & 0.59 & 0.76 & 0.55 \\
        SAM-2 Fine-tuned & {\color{green}\cmark} & 6.05 & \textbf{0.72} & \textbf{0.76} & \textbf{0.78} & \textbf{0.90} & \textbf{0.77} & \textbf{0.76} & \textbf{0.90} & \textbf{0.74} & \textbf{0.83} & \textbf{0.93} & \textbf{0.82} \\
        \bottomrule	 
    \end{tabular}
    } 
    \vspace{-0.5em}
    \caption{Performance of benchmarking repurposed models for hierarchical segmentation (top), VIS (middle), VOS (bottom) on our dataset. (\textbf{Inf./Vid.} =  Average number of inference passes per video)}
    \label{tab:HIT_model_results}
\end{table*}

\vspace{-1em}\paragraph{Evaluation Metrics.}
We evaluate with respect to five metrics. Two are the metrics we introduced for our hierarchical instance tracking task: MOTA-H and MOTA-OBJ. The other three provide backward compatibility for analysis with video object segmentation (VOS), video instance segmentation (VIS), and hierarchical segmentation methods, but are all image-based (i.e., \emph{ignore tracking}). First is $J \& F$~\cite{pont20172017}, the standard metric for VOS, which computes the mean between the Jaccard Index ($J$) (i.e., aka, intersection over union) and the boundary F-measure ($F$), the harmonic mean of precision and recall .  The next two are average precision ($AP$) and average recall ($AR$), the standard metrics for VIS. Hierarchical segmentation papers~\cite{wang2024hierarchical, ding2023VDT, li2023} also use $AP$ scores, with the key distinction from VIS methods that they report scores for both object and part categories (separately).  The final three metrics all result in scores that range from 0 to 1, with larger scores (i.e., closer to 1) signifying better performance.

\subsection{Hierarchical Image Segmentation}
\label{sec:hier-seg}
One relevant family of models for our task are those performing hierarchical instance segmentation.  That is because they can provide the first critical step for tracking of locating all relevant objects and parts in each frame, and then leave it up to downstream association methods (e.g., Hungarian matching) to match segmentation masks across video frames to create masklets.  

\vspace{-1em}\paragraph{Model.}
We evaluate the top-performing hierarchical image segmentation model called Hierarchical Open-vocabulary Universal Image Segmentation (HIPIE)~\cite{wang2024hierarchical}.  As noted in its name, the model is designed to support any vocabulary without further training.  It outputs which categories are present where in a given image, when provided as input a list of all candidate categories that it should find.  We feed the model as input our 40 object and part categories.  We test two publicly-available variants, which rely on different backbones: ResNet-50 \cite{he2016deep} and ViT-H \cite{dosovitskiy2020image}.

\vspace{-1em}\paragraph{Results.}
Results are shown in the first two rows of \textbf{Table~\ref{tab:HIT_model_results}}.  Both variants failed completely, indicating no categories were present for nearly all the images where at least one relevant category was actually present. We attribute HIPIE's poor performance to poor generalization abilities, despite its claim to support an open vocabulary. Thus, we conclude current hierarchical image segmentation models are an inadequate foundation to extend for tracking, as they provide no detected instances for downstream associate methods to associate across frames.

\begin{figure*}[t!]
    \centering
    \includegraphics[width=1.0\textwidth]{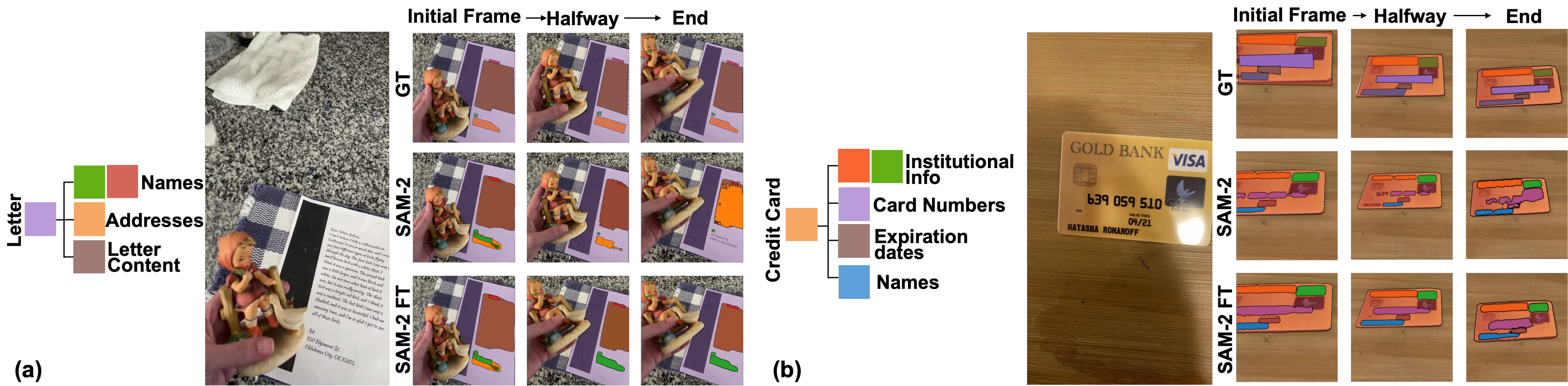}
    \vspace{-2em}
    \caption{Initial video frames followed by cropped ground truths and predictions from SAM-2, as is and fine-tuned, in subsequent frames.}
    \label{fig:qualitative-results}
\end{figure*}

\subsection{Video Instance Segmentation}
Another relevant family of models are those for video instance segmentation (VIS), which track specified semantic categories.  While they cannot support our task in a single inference pass, since they do not permit the same pixel to belong to multiple semantic categories, a workaround is to instead develop two VIS models developed to support objects and parts separately and then leverage post-processing to associate parts with parent objects.

\vspace{-1em}\paragraph{Model.}
We benchmark two variants of the popular, top-performing model called Mask2Former \cite{mask2former}.  One variant results from fine-tuning two models for object and parts respectively, and then associating all masklets for parts to the masklet for the detected object, since a bias of our videos is they show only a single tracked object.  The other variant results from fine-tuning a single Mask2Former to simultaneously support all object and part categories in our dataset. 

\vspace{-1em}\paragraph{Results.}
Results are shown in the second and third rows of \textbf{Table~\ref{tab:HIT_model_results}}.  Overall, both methods perform poorly with respect to both the tracking-based and segmentation-based metrics.  Moreover, both Mask2Formers performed worse for parts than objects, with the single Mask2Former trained to support both types neglecting parts entirely (i.e., scores of 0 for part-only metrics) in order to instead prioritize finding objects.  We suspect the bias towards objects stem from a combination of this prioritization during initial training as well as that objects can be easier to locate, possibly due to their larger sizes and greater contrast of appearance relative to surrounding content.

Nonetheless, we observe a considerable performance boost from these models compared to the hierarchical instance segmentation model.  We attribute this largely to Mask2Former being trained on in-domain data reflective of the test set.  Moreover, it is promising to see that Mask2Former trained on only parts can succeed at times; e.g., AR of 0.32 and AP of 0.11.  A valuable direction for future work is to explore if greater boosts can be secured from more training data, or if fundamentally new architectures are needed.

\subsection{Video Object Segmentation}
Last, we evaluate video object segmentation (VOS) models. Like VIS models, they require multiple inference passes to recover objects and parts.  Unlike VIS models, this is because VOS models only track one entity at a time, necessitating an inference pass for each object and part alongside post-processing to associate parts with objects.  VOS models also require as input the target entity's segmentation in the first frame it appears.

\vspace{-1em}\paragraph{Models.}
We evaluate two popular VOS models---SAM-2~\cite{ravi2024sam} and XMem++~\cite{xmem}---designed to track anything.  Both are configured to \emph{cheat}, receiving ground truth segmentation of each target entity in the first frame that each appears so they track entities only after knowing where to look when.  This is necessary because automated localization from a model is not yet suitable (Section~\ref{sec:hier-seg}).  

\vspace{-1em}\paragraph{Results.}
Results in \textbf{Table~\ref{tab:HIT_model_results}} (rows 5-6) show these models outperform all other benchmarked models across metrics. This underscores a strong benefit of automating the cheating manual annotation step of locating target categories in images.  This highlights a clear direction for future work: automating the manual annotation step we introduced, which currently provides the model with ground-truth part locations and artificially inflates performance. 

Further analysis indicates that while VOS models often segment parts correctly in individual video frames, they struggle to consistently associate them across frames, leading to identity switching.  This is reflect in high image-based segmentation scores (i.e., all exceed 0.5) alongside comparatively low tracking scores (i.e., MOTA-H below 0.5), suggesting that future work should also focus on improving mechanisms for associating part masks across frames.

Reinforcing earlier observations, a substantial performance gap remains between parts and objects.  Suspecting this stems from a domain shift between SAM-2's original training data and our dataset, we fine-tuned SAM-2 on our training split.\footnote{XMem++ wasn't fine-tuned due to no public scripts or documentation.}  The results in the final row of \textbf{Table~\ref{tab:HIT_model_results}} show a considerable improvement, particularly in MOTA-H, which nearly doubles from 0.39 to 0.72.  We attribute this to SAM-2 learning a more effective appearance model for text-based parts, improving mask association across neighboring frames.  Qualitatively, \textbf{Figure~\ref{fig:qualitative-results}} shows the left sequence resolving identity switches between the ``address" and ``letter content" categories and the right sequence producing more uniform masks that exclude irrelevant regions (e.g. for the card number). While SAM-2 is designed to track \emph{anything}, these results highlight that specialized datasets like BIV-Priv-HIT can substantially improve its tracking ability. 
\section{Conclusions}
\label{sec:conclusions}
We introduce the novel task of hierarchical instance tracking along with the first dataset designed to support this task, called BIV-Priv-HIT. We also introduce an evaluation metric, MOTA-H to benchmark models' performance for the task. Our analysis reveals the unique characteristics of BIV-Priv-HIT compared to existing datasets. Benchmarking modern video object segmentation, video instance segmentation, and hierarchical image segmentation models demonstrate the dataset provides a challenging problem for the research community.  

While this work advances hierarchical instance tracking, it has limitations.  For example MOTA-H is based on MOTA, which can conflate different error types; other metrics (e.g., HOTA~\cite{luiten2021hota}, TETA~\cite{li2022tracking}) may provide a clearer assessment. The dataset also focuses on a limited set of part categories and domain, which may restrict generalization to other real-world scenarios. This work also carries ethical risks from misuse of models developed using the dataset, which future work should mitigate through developing responsible-use safeguards.

\vspace{-1em}\paragraph{Acknowledgments.}
We thank Josh Myers-Dean for his assistance with setting up the models.  This project was supported by National Science Foundation SaTC awards (\#2148080, \#2126314, and \#2125925).

{
    \small
    \bibliographystyle{ieeenat_fullname}
    \bibliography{main}

@article{li2025exploring,
  title={Exploring Object Status Recognition for Recipe Progress Tracking in Non-Visual Cooking},
  author={Li, Franklin Mingzhe and Ng, Kaitlyn and Zhu, Bin and Carrington, Patrick},
  journal={arXiv preprint arXiv:2507.03330},
  year={2025}
}

@inproceedings{kim2019korean,
  title={Korean localization of visual question answering for blind people},
  author={Kim, Jin-Hwa and Lim, Soohyun and Park, Jaesun and Cho, Hansu},
  booktitle={SK T-Brain-AI for Social Good Workshop at NeurIPS},
  volume={2},
  year={2019}
}

@misc{vis2022,
  author       = "Linjie Yang and Yuchen Fan and Ning Xu",
  title        = "The 4th Large-scale Video Object Segmentation Challenge - video instance segmentation track",
  month        = jun,
  year         = "2022",
}

@inproceedings{ramanathan2023paco,
  title={Paco: Parts and attributes of common objects},
  author={Ramanathan, Vignesh and Kalia, Anmol and Petrovic, Vladan and Wen, Yi and Zheng, Baixue and Guo, Baishan and Wang, Rui and Marquez, Aaron and Kovvuri, Rama and Kadian, Abhishek and others},
  booktitle={Proceedings of the IEEE/CVF Conference on Computer Vision and Pattern Recognition},
  pages={7141--7151},
  year={2023}
}

@article{zhou2019semantic,
  title={Semantic understanding of scenes through the ade20k dataset},
  author={Zhou, Bolei and Zhao, Hang and Puig, Xavier and Xiao, Tete and Fidler, Sanja and Barriuso, Adela and Torralba, Antonio},
  journal={International Journal of Computer Vision},
  volume={127},
  pages={302--321},
  year={2019},
  publisher={Springer}
}

@inproceedings{he2022partimagenet,
  title={Partimagenet: A large, high-quality dataset of parts},
  author={He, Ju and Yang, Shuo and Yang, Shaokang and Kortylewski, Adam and Yuan, Xiaoding and Chen, Jie-Neng and Liu, Shuai and Yang, Cheng and Yu, Qihang and Yuille, Alan},
  booktitle={European Conference on Computer Vision},
  pages={128--145},
  year={2022},
  organization={Springer}
}

@inproceedings{chen2014detect,
  title={Detect what you can: Detecting and representing objects using holistic models and body parts},
  author={Chen, Xianjie and Mottaghi, Roozbeh and Liu, Xiaobai and Fidler, Sanja and Urtasun, Raquel and Yuille, Alan},
  booktitle={Proceedings of the IEEE conference on computer vision and pattern recognition},
  pages={1971--1978},
  year={2014}
}

@inproceedings{de2021part,
  title={Part-aware panoptic segmentation},
  author={de Geus, Daan and Meletis, Panagiotis and Lu, Chenyang and Wen, Xiaoxiao and Dubbelman, Gijs},
  booktitle={Proceedings of the IEEE/CVF Conference on Computer Vision and Pattern Recognition},
  pages={5485--5494},
  year={2021}
}

@inproceedings{lin2014microsoft,
  title={Microsoft coco: Common objects in context},
  author={Lin, Tsung-Yi and Maire, Michael and Belongie, Serge and Hays, James and Perona, Pietro and Ramanan, Deva and Doll{\'a}r, Piotr and Zitnick, C Lawrence},
  booktitle={Computer Vision--ECCV 2014: 13th European Conference, Zurich, Switzerland, September 6-12, 2014, Proceedings, Part V 13},
  pages={740--755},
  year={2014},
  organization={Springer}
}

@inproceedings{sharma2023disability,
  title={Disability-first design and creation of a dataset showing private visual information collected with people who are blind},
  author={Sharma, Tanusree and Stangl, Abigale and Zhang, Lotus and Tseng, Yu-Yun and Xu, Inan and Findlater, Leah and Gurari, Danna and Wang, Yang},
  booktitle={Proceedings of the 2023 CHI Conference on Human Factors in Computing Systems},
  pages={1--15},
  year={2023}
}

@inproceedings{reynolds2024salient,
  title={Salient object detection for images taken by people with vision impairments},
  author={Reynolds, Jarek and Nagesh, Chandra Kanth and Gurari, Danna},
  booktitle={Proceedings of the IEEE/CVF Winter Conference on Applications of Computer Vision},
  pages={8522--8531},
  year={2024}
}

@article{tseng2024biv,
  title={BIV-Priv-Seg: Locating Private Content in Images Taken by People With Visual Impairments},
  author={Tseng, Yu-Yun and Sharma, Tanusree and Zhang, Lotus and Stangl, Abigale and Findlater, Leah and Wang, Yang and Tseng, Danna Gurari Yu-Yun and Gurari, Danna},
  journal={arXiv preprint arXiv:2407.18243},
  year={2024}
}

@article{ravi2024sam,
  title={Sam 2: Segment anything in images and videos},
  author={Ravi, Nikhila and Gabeur, Valentin and Hu, Yuan-Ting and Hu, Ronghang and Ryali, Chaitanya and Ma, Tengyu and Khedr, Haitham and R{\"a}dle, Roman and Rolland, Chloe and Gustafson, Laura and others},
  journal={arXiv preprint arXiv:2408.00714},
  year={2024}
}

@misc{vos2022,
  author       = "Linjie Yang and Yuchen Fan and Ning Xu",
  title        = "The 4th Large-scale Video Object Segmentation Challenge - video object segmentation track",
  month        = jun,
  year         = "2022",
}

@article{pont20172017,
  title={The 2017 davis challenge on video object segmentation},
  author={Pont-Tuset, Jordi and Perazzi, Federico and Caelles, Sergi and Arbel{\'a}ez, Pablo and Sorkine-Hornung, Alex and Van Gool, Luc},
  journal={arXiv preprint arXiv:1704.00675},
  year={2017}
}

@inproceedings{tseng2022vizwiz,
  title={Vizwiz-fewshot: Locating objects in images taken by people with visual impairments},
  author={Tseng, Yu-Yun and Bell, Alexander and Gurari, Danna},
  booktitle={European Conference on Computer Vision},
  pages={575--591},
  year={2022},
  organization={Springer}
}

@inproceedings{chen2022grounding,
  title={Grounding answers for visual questions asked by visually impaired people},
  author={Chen, Chongyan and Anjum, Samreen and Gurari, Danna},
  booktitle={Proceedings of the IEEE/CVF Conference on Computer Vision and Pattern Recognition},
  pages={19098--19107},
  year={2022}
}

@inproceedings{gurari2019vizwiz,
  title={Vizwiz-priv: A dataset for recognizing the presence and purpose of private visual information in images taken by blind people},
  author={Gurari, Danna and Li, Qing and Lin, Chi and Zhao, Yinan and Guo, Anhong and Stangl, Abigale and Bigham, Jeffrey P},
  booktitle={Proceedings of the IEEE/CVF Conference on Computer Vision and Pattern Recognition},
  pages={939--948},
  year={2019}
}

@inproceedings{li2020towards,
  title={Towards a taxonomy of content sensitivity and sharing preferences for photos},
  author={Li, Yifang and Vishwamitra, Nishant and Hu, Hongxin and Caine, Kelly},
  booktitle={Proceedings of the 2020 CHI Conference on Human Factors in Computing Systems},
  pages={1--14},
  year={2020}
}

@misc{PII_Glossary, title={PII - Glossary: CSRC}, url={https://csrc.nist.gov/glossary/term/PII#:~:text=Personally%20Identifiable%20Information%3B%20Any%20representation,800%2D79%2D2%20from%20EGovAct}, journal={CSRC Content Editor}, author={Editor, CSRC Content}}

@article{pei2023tale,
  title={A tale of two communities: Privacy of third party app users in crowdsourcing-the case of receipt transcription},
  author={Pei, Weiping and Likhtenshteyn, Yanina and Yue, Chuan},
  journal={Proceedings of the ACM on Human-Computer Interaction},
  volume={7},
  number={CSCW2},
  pages={1--43},
  year={2023},
  publisher={ACM New York, NY, USA}
}

@article{xu2024dipa2,
  title={DIPA2: An Image Dataset with Cross-cultural Privacy Perception Annotations},
  author={Xu, Anran and Zhou, Zhongyi and Miyazaki, Kakeru and Yoshikawa, Ryo and Hosio, Simo and Yatani, Koji},
  journal={Proceedings of the ACM on Interactive, Mobile, Wearable and Ubiquitous Technologies},
  volume={7},
  number={4},
  pages={1--30},
  year={2024},
  publisher={ACM New York, NY, USA}
}

@inproceedings{zhou2017scene,
  title={Scene parsing through ade20k dataset},
  author={Zhou, Bolei and Zhao, Hang and Puig, Xavier and Fidler, Sanja and Barriuso, Adela and Torralba, Antonio},
  booktitle={Proceedings of the IEEE conference on computer vision and pattern recognition},
  pages={633--641},
  year={2017}
}

@article{liang2018look,
  title={Look into person: Joint body parsing \& pose estimation network and a new benchmark},
  author={Liang, Xiaodan and Gong, Ke and Shen, Xiaohui and Lin, Liang},
  journal={IEEE transactions on pattern analysis and machine intelligence},
  volume={41},
  number={4},
  pages={871--885},
  year={2018},
  publisher={IEEE}
}

@inproceedings{zhao2018understanding,
  title={Understanding humans in crowded scenes: Deep nested adversarial learning and a new benchmark for multi-human parsing},
  author={Zhao, Jian and Li, Jianshu and Cheng, Yu and Sim, Terence and Yan, Shuicheng and Feng, Jiashi},
  booktitle={Proceedings of the 26th ACM international conference on Multimedia},
  pages={792--800},
  year={2018}
}

@inproceedings{gong2018instance,
  title={Instance-level human parsing via part grouping network},
  author={Gong, Ke and Liang, Xiaodan and Li, Yicheng and Chen, Yimin and Yang, Ming and Lin, Liang},
  booktitle={Proceedings of the European conference on computer vision (ECCV)},
  pages={770--785},
  year={2018}
}

@article{wah2011caltech,
  title={The caltech-ucsd birds-200-2011 dataset},
  author={Wah, Catherine and Branson, Steve and Welinder, Peter and Perona, Pietro and Belongie, Serge},
  year={2011},
  publisher={California Institute of Technology}
}

@inproceedings{zheng2018modanet,
  title={Modanet: A large-scale street fashion dataset with polygon annotations},
  author={Zheng, Shuai and Yang, Fan and Kiapour, M Hadi and Piramuthu, Robinson},
  booktitle={Proceedings of the 26th ACM international conference on Multimedia},
  pages={1670--1678},
  year={2018}
}

@inproceedings{song2019apollocar3d,
  title={Apollocar3d: A large 3d car instance understanding benchmark for autonomous driving},
  author={Song, Xibin and Wang, Peng and Zhou, Dingfu and Zhu, Rui and Guan, Chenye and Dai, Yuchao and Su, Hao and Li, Hongdong and Yang, Ruigang},
  booktitle={Proceedings of the IEEE/CVF Conference on Computer Vision and Pattern Recognition},
  pages={5452--5462},
  year={2019}
}

@inproceedings{jia2020fashionpedia,
  title={Fashionpedia: Ontology, segmentation, and an attribute localization dataset},
  author={Jia, Menglin and Shi, Mengyun and Sirotenko, Mikhail and Cui, Yin and Cardie, Claire and Hariharan, Bharath and Adam, Hartwig and Belongie, Serge},
  booktitle={Computer Vision--ECCV 2020: 16th European Conference, Glasgow, UK, August 23--28, 2020, Proceedings, Part I 16},
  pages={316--332},
  year={2020},
  organization={Springer}
}

@article{wei2024ov,
  title={Ov-parts: Towards open-vocabulary part segmentation},
  author={Wei, Meng and Yue, Xiaoyu and Zhang, Wenwei and Kong, Shu and Liu, Xihui and Pang, Jiangmiao},
  journal={Advances in Neural Information Processing Systems},
  volume={36},
  year={2024}
}

@article{wang2024hierarchical,
  title={Hierarchical open-vocabulary universal image segmentation},
  author={Wang, Xudong and Li, Shufan and Kallidromitis, Konstantinos and Kato, Yusuke and Kozuka, Kazuki and Darrell, Trevor},
  journal={Advances in Neural Information Processing Systems},
  volume={36},
  year={2024}
}

@inproceedings{he2016deep,
  title={Deep residual learning for image recognition},
  author={He, Kaiming and Zhang, Xiangyu and Ren, Shaoqing and Sun, Jian},
  booktitle={Proceedings of the IEEE conference on computer vision and pattern recognition},
  pages={770--778},
  year={2016}
}

@article{dosovitskiy2020image,
  title={An image is worth 16x16 words: Transformers for image recognition at scale},
  author={Dosovitskiy, Alexey},
  journal={arXiv preprint arXiv:2010.11929},
  year={2020}
}

@inproceedings{zhang2024designing,
  title={Designing Accessible Obfuscation Support for Blind Individuals’ Visual Privacy Management},
  author={Zhang, Lotus and Stangl, Abigale and Sharma, Tanusree and Tseng, Yu-Yun and Xu, Inan and Gurari, Danna and Wang, Yang and Findlater, Leah},
  booktitle={Proceedings of the CHI Conference on Human Factors in Computing Systems},
  pages={1--19},
  year={2024}
}

@article{samson2024privacy,
  title={Privacy-Aware Visual Language Models},
  author={Samson, Laurens and Barazani, Nimrod and Ghebreab, Sennay and Asano, Yuki M},
  journal={arXiv preprint arXiv:2405.17423},
  year={2024}
}

@inproceedings{Yang2019vis,
    author = {Linjie Yang and Yuchen Fan and Ning Xu},
    title = {Video instance segmentation},
    booktitle = {ICCV},
    year = {2019},
 }

@misc{orbit,
  doi = {10.25383/CITY.14294597},
  
  url = {https://city.figshare.com/articles/dataset/ORBIT_A_real-world_few-shot_dataset_for_teachable_object_recognition_collected_from_people_who_are_blind_or_low_vision/14294597},
  
  author = {Massiceti, Daniela and Theodorou, Lida and Zintgraf, Luisa and Harris, Matthew Tobias and Stumpf, Simone and Morrison, Cecily and Cutrell, Edward and Hofmann, Katja},
  
  title = {ORBIT: A real-world few-shot dataset for teachable object recognition collected from people who are blind or low vision},
  
  publisher = {City, University of London},
  
  year = {2021},
  
  copyright = {Creative Commons Attribution 4.0 International}
}

@misc{xmem,
      title={XMem++: Production-level Video Segmentation From Few Annotated Frames}, 
      author={Maksym Bekuzarov and Ariana Bermudez and Joon-Young Lee and Hao Li},
      year={2023},
      eprint={2307.15958},
      archivePrefix={arXiv},
      primaryClass={cs.CV},
      url={https://arxiv.org/abs/2307.15958}, 
}

@misc{mask2former,
      title={Mask2Former for Video Instance Segmentation}, 
      author={Bowen Cheng and Anwesa Choudhuri and Ishan Misra and Alexander Kirillov and Rohit Girdhar and Alexander G. Schwing},
      year={2021},
      eprint={2112.10764},
      archivePrefix={arXiv},
      primaryClass={cs.CV},
      url={https://arxiv.org/abs/2112.10764}, 
}

@misc{gurari2018,
      title={VizWiz Grand Challenge: Answering Visual Questions from Blind People}, 
      author={Danna Gurari and Qing Li and Abigale J. Stangl and Anhong Guo and Chi Lin and Kristen Grauman and Jiebo Luo and Jeffrey P. Bigham},
      year={2018},
      eprint={1802.08218},
      archivePrefix={arXiv},
      primaryClass={cs.CV},
      url={https://arxiv.org/abs/1802.08218}, 
}

@misc{gurari2020,
      title={Captioning Images Taken by People Who Are Blind}, 
      author={Danna Gurari and Yinan Zhao and Meng Zhang and Nilavra Bhattacharya},
      year={2020},
      eprint={2002.08565},
      archivePrefix={arXiv},
      primaryClass={cs.CV},
      url={https://arxiv.org/abs/2002.08565}, 
}

@misc{EgoBlind,
      title={EgoBlind: Towards Egocentric Visual Assistance for the Blind}, 
      author={Junbin Xiao and Nanxin Huang and Hao Qiu and Zhulin Tao and Xun Yang and Richang Hong and Meng Wang and Angela Yao},
      year={2025},
      eprint={2503.08221},
      archivePrefix={arXiv},
      primaryClass={cs.CV},
      url={https://arxiv.org/abs/2503.08221}, 
}

@misc{BLVnavigation,
      title={A Dataset for Crucial Object Recognition in Blind and Low-Vision Individuals' Navigation}, 
      author={Md Touhidul Islam and Imran Kabir and Elena Ariel Pearce and Md Alimoor Reza and Syed Masum Billah},
      year={2024},
      eprint={2407.16777},
      archivePrefix={arXiv},
      primaryClass={cs.CV},
      url={https://arxiv.org/abs/2407.16777}, 
}

@misc{ding2023VDT,
      title={Visual Dependency Transformers: Dependency Tree Emerges from Reversed Attention}, 
      author={Mingyu Ding and Yikang Shen and Lijie Fan and Zhenfang Chen and Zitian Chen and Ping Luo and Joshua B. Tenenbaum and Chuang Gan},
      year={2023},
      eprint={2304.03282},
      archivePrefix={arXiv},
      primaryClass={cs.CV},
      url={https://arxiv.org/abs/2304.03282}, 
}

@misc{li2023,
      title={Semantic-SAM: Segment and Recognize Anything at Any Granularity}, 
      author={Feng Li and Hao Zhang and Peize Sun and Xueyan Zou and Shilong Liu and Jianwei Yang and Chunyuan Li and Lei Zhang and Jianfeng Gao},
      year={2023},
      eprint={2307.04767},
      archivePrefix={arXiv},
      primaryClass={cs.CV},
      url={https://arxiv.org/abs/2307.04767}, 
}

@article{MOTA,
  title={Evaluating Multiple Object Tracking Performance: The CLEAR MOT Metrics},
  author={K. Bernardin and R. Stiefelhagen},
  journal={Journal of Image and Video Processing},
  year={2008},
  volume={2008},
  pages={246309:1-246309:10}
}

@misc{lukežič2016,
title={Deformable Parts Correlation Filters for Robust Visual Tracking},
author={Alan Lukežič and Luka Čehovin and Matej Kristan},
year={2016},
eprint={1605.03720},
archivePrefix={arXiv},
primaryClass={[cs.CV](http://cs.cv/)},
url={https://arxiv.org/abs/1605.03720},
}

@article{akin2016,
title = {Deformable part-based tracking by coupled global and local correlation filters},
journal = {Journal of Visual Communication and Image Representation},
volume = {38},
pages = {763-774},
year = {2016},
issn = {1047-3203},
doi = {https://doi.org/10.1016/j.jvcir.2016.04.018},
url = {https://www.sciencedirect.com/science/article/pii/S1047320316300517},
author = {Osman Akin and Erkut Erdem and Aykut Erdem and Krystian Mikolajczyk},
}

@ARTICLE{Yao2017,
  author={Yao, Rui and Shi, Qinfeng and Shen, Chunhua and Zhang, Yanning and van den Hengel, Anton},
  journal={IEEE Transactions on Circuits and Systems for Video Technology},
  title={Part-Based Robust Tracking Using Online Latent Structured Learning},
  year={2017},
  volume={27},
  number={6},
  pages={1235-1248},
  doi={10.1109/TCSVT.2016.2527358}}

@misc{cheng2018,
title={Fast and Accurate Online Video Object Segmentation via Tracking Parts},
author={Jingchun Cheng and Yi-Hsuan Tsai and Wei-Chih Hung and Shengjin Wang and Ming-Hsuan Yang},
year={2018},
eprint={1806.02323},
archivePrefix={arXiv},
primaryClass={[cs.CV](http://cs.cv/)},
url={https://arxiv.org/abs/1806.02323},
}

@article{de2018part,
  title={Part-based tracking by sampling},
  author={De Ath, George and Everson, Richard M},
  journal={arXiv preprint arXiv:1805.08511},
  year={2018}
}

@inproceedings{yao2013part,
  title={Part-based visual tracking with online latent structural learning},
  author={Yao, Rui and Shi, Qinfeng and Shen, Chunhua and Zhang, Yanning and Van Den Hengel, Anton},
  booktitle={Proceedings of the IEEE conference on computer vision and pattern recognition},
  pages={2363--2370},
  year={2013}
}

@inproceedings{zhang2023multiple,
  title={Multiple planar object tracking},
  author={Zhang, Zhicheng and Liu, Shengzhe and Yang, Jufeng},
  booktitle={Proceedings of the IEEE/CVF International Conference on Computer Vision},
  pages={23460--23470},
  year={2023}
}

@article{huang2017visual,
  title={Visual tracking by sampling in part space},
  author={Huang, Lianghua and Ma, Bo and Shen, Jianbing and He, Hui and Shao, Ling and Porikli, Fatih},
  journal={IEEE Transactions on Image Processing},
  volume={26},
  number={12},
  pages={5800--5810},
  year={2017},
  publisher={IEEE}
}

@inproceedings{burceanu2018learning,
  title={Learning a robust society of tracking parts using co-occurrence constraints},
  author={Burceanu, Elena and Leordeanu, Marius},
  booktitle={Proceedings of the European Conference on Computer Vision (ECCV) Workshops},
  pages={0--0},
  year={2018}
}

@inproceedings{bhattacharya2019does,
  title={Why does a visual question have different answers?},
  author={Bhattacharya, Nilavra and Li, Qing and Gurari, Danna},
  booktitle={Proceedings of the IEEE/CVF international conference on computer vision},
  pages={4271--4280},
  year={2019}
}

@inproceedings{chiu2020assessing,
  title={Assessing image quality issues for real-world problems. In 2020 IEEE},
  author={Chiu, Tai-Yin and Zhao, Yinan and Gurari, Danna},
  booktitle={CVF Conference on Computer Vision and Pattern Recognition (CVPR)},
  pages={3643--3653},
  year={2020}
}

@article{zeng2020vision,
  title={Vision skills needed to answer visual questions},
  author={Zeng, Xiaoyu and Wang, Yanan and Chiu, Tai-Yin and Bhattacharya, Nilavra and Gurari, Danna},
  journal={Proceedings of the ACM on Human-Computer Interaction},
  volume={4},
  number={CSCW2},
  pages={1--31},
  year={2020},
  publisher={ACM New York, NY, USA}
}

@inproceedings{bafghi2023new,
  title={A new dataset based on images taken by blind people for testing the robustness of image classification models trained for imagenet categories},
  author={Bafghi, Reza Akbarian and Gurari, Danna},
  booktitle={Proceedings of the IEEE/CVF Conference on Computer Vision and Pattern Recognition},
  pages={16261--16270},
  year={2023}
}

@inproceedings{chen2023vqa,
  title={Vqa therapy: Exploring answer differences by visually grounding answers},
  author={Chen, Chongyan and Anjum, Samreen and Gurari, Danna},
  booktitle={Proceedings of the IEEE/CVF International Conference on Computer Vision},
  pages={15315--15325},
  year={2023}
}

@inproceedings{huh2024long,
  title={Long-form answers to visual questions from blind and low vision people},
  author={Huh, Mina and Xu, Fangyuan and Peng, Yi-Hao and Chen, Chongyan and Gurari, Danna and Choi, Eunsol and Pavel, Amy},
  booktitle={Workshop on Demographic Diversity in Computer Vision@ CVPR 2025},
  year={2024}
}

@inproceedings{chen2025acknowledging,
  title={Acknowledging Focus Ambiguity in Visual Questions},
  author={Chen, Chongyan and Tseng, Yu-Yun and Li, Zhuoheng and Venkatesh, Anush and Gurari, Danna},
  booktitle={Proceedings of the IEEE/CVF International Conference on Computer Vision},
  pages={1228--1238},
  year={2025}
}

@article{luiten2021hota,
  title={Hota: A higher order metric for evaluating multi-object tracking},
  author={Luiten, Jonathon and Osep, Aljosa and Dendorfer, Patrick and Torr, Philip and Geiger, Andreas and Leal-Taix{\'e}, Laura and Leibe, Bastian},
  journal={International journal of computer vision},
  volume={129},
  number={2},
  pages={548--578},
  year={2021},
  publisher={Springer}
}

@inproceedings{li2022tracking,
  title={Tracking every thing in the wild},
  author={Li, Siyuan and Danelljan, Martin and Ding, Henghui and Huang, Thomas E and Yu, Fisher},
  booktitle={European conference on computer vision},
  pages={498--515},
  year={2022},
  organization={Springer}
}
}

\clearpage
\pagebreak
\noindent {\LARGE \textbf{Supplementary Materials}}\\
\noindent
\\
This document supplements the main paper with additional information concerning:

\begin{enumerate}[label=\Alph*.]
    \item Dataset Creation (supplements Section 3.1)
    \begin{itemize}
        \item Video Source 
        \item Annotation Collection
        \item Ground Truth Generation 
    \end{itemize}
    \item Dataset Analysis (supplements Section 3.2)
    \begin{itemize}
        \item Baseline Datasets for Comparison
        \item Dataset Composition 
        \item Segmentation Properties 
    \end{itemize}
    \item Model Benchmarking (supplements Section  4)
    \begin{itemize}
        \item Fine-Grained Analysis 
    \end{itemize}
\end{enumerate}

\renewcommand{\thesection}{\Alph{section}}
\setcounter{section}{0}

\section{Dataset Creation}

\paragraph{Video Source.} 
As noted in the main paper, two in-house annotators specified for each of the 552 videos the start and end frames when objects of interest where visible. We employed the Intersection Over Union (IoU) similarity score to gauge similarity among the annotator-flagged start and stop frames. For the intersection, we calculated the duration between the maximum value of the two annotated start times and the minimum value of the two annotated end times. For the union, we calculated the duration between the minimum value of the two annotated start times and the maximum value of the two annotated end times. We used an IoU threshold of 0.99 to determine whether the start and end frame annotations match. 

\vspace{-0.5em}\paragraph{Annotation Collection.}
We hired crowdworkers on Amazon Mechanical Turk to annotate our objects and parts with an annotation interface that we built. The interface  collects a series of clicked points to create connected polygons on independent video frames. The interface supports annotating multiple polygons to capture when (1) there are multiple instances of a part (e.g., multiple account numbers) and (2) occlusions that break a part's appearance into multiple, disconnected pieces. Workers were given a comprehensive instruction set including instructions on how to segment each object class along with its parts. 

To facilitate collection of high-quality annotations, we employed several quality control checks. We monitored ongoing quality by reviewing outliers regarding worker's frequency of indicating object and part non-presence, average time to complete a full annotation task, and the level of detail they provided in their segmentations (e.g., high prevalence of triangles).  We conducted manual spot-checks at the conclusion of each phased task rollout.

\vspace{-0.5em}\paragraph{Ground Truth Generation.}
We used redundant annotations to establish ground truth for objects.  

We observed annotation agreement regarding the presence of an object for 96.5\% of frames (present in 9,804 frames and absent in 971 frames), with 93\% of the remaining 361 frames showing the object. Consequently, 91\% (10,165) of the 11,165 annotated frames showed a target object.  Of these, 98\% (9980) were similar while 2\% (185) had IoU scores less than 0.75 or lacked a redundant annotation necessary to calculate an IoU similarity score.  For those lacking annotation agreement, the in-house annotators reviewed both annotations side by side and then chose one of the two annotations to keep for ground truth for 95\% (175) of instances and resegmented the other 4\% (10) where an object was missing or misidentified. 

We observed annotation agreement that parts were not present for 43\% (19,201) of 44,600 instances where crowdworkers were prompted about a part's presence.  Of the parts deemed present, 67.8\% (17,217) had high segmentation similarity and the remaining 32.2\% (8,182) went through further manual review.  An in-house annotator reviewed both part-level annotations and then selected the correct option when available or created a new segmentation when neither were suitable. Of the 8,182 part-level annotations, one part-level annotation was selected for 53.2\% (4,357) instances and new segmentations were created for the rest.

\begin{figure*}[t!]
    \centering
    \includegraphics[width=\textwidth]{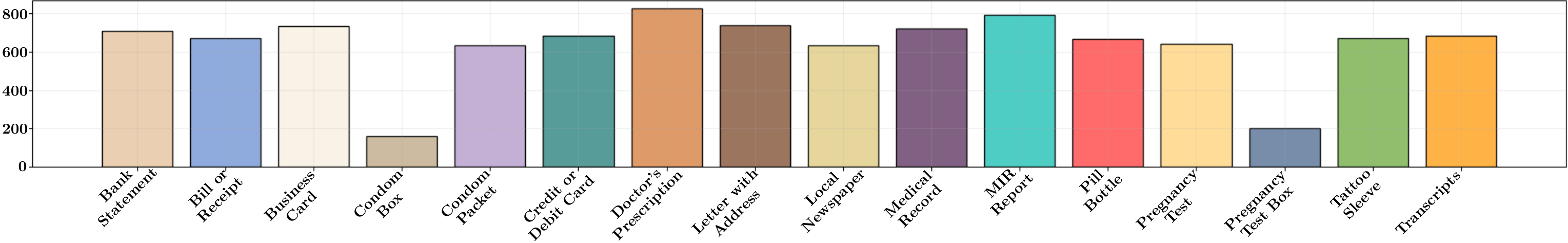}
    \vspace{-1.5em}
    \caption{BIV-Priv-HIT object annotation frequency distribution of objects across all 10,165 object annotations}
     \label{fig:obj-annots-per-obj}
\end{figure*}

\begin{figure*}[t!]
    \centering
    \includegraphics[width=\textwidth]{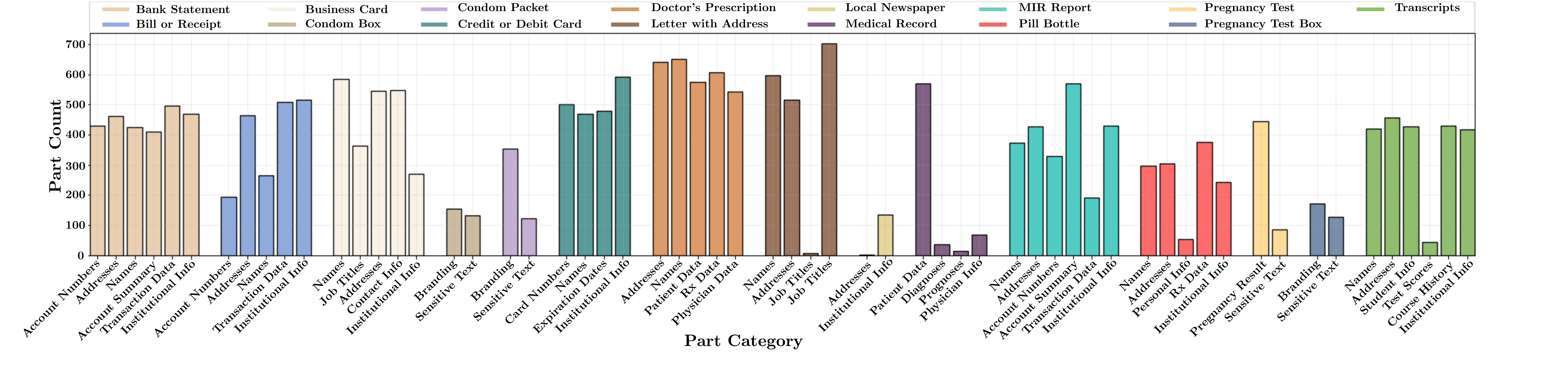}
    \vspace{-1.5em}
    \caption{BIV-Priv-HIT part annotation frequency distribution of part categories across all object 10,165 annotations.}
     \label{fig:part-annots-per-obj}
\end{figure*}

\section{Dataset Analysis}

\paragraph{Baseline Datasets for Comparison.}
Only one other dataset could feasibly support hierarchically tracking objects and parts, Meta's SA-V~\cite{ravi2024sam}, since it provides both object and part masklets.  However, it is non-trivial to determine the hierarchical object-part relations automatically.  Specifically, inference is necessary because part and object masklets are treated the same, yet this is non-trivial to achieve for numerous reasons including that unrelated occluding entities can lead one to incorrectly deem an entity to be a part (e.g., a watch on a person's wrist). 

\vspace{-0.5em}\paragraph{Dataset Composition.}
The object category frequency distribution across the BIV-Priv-HIT dataset is shown in \textbf{Figure~\ref{fig:obj-annots-per-obj}}. When observing the object category frequency distribution, the condom and pregnancy test boxes have the lowest object counts. This is because, in the original dataset, they are both categorized under the same label, `condom' and `pregnancy test.' In contrast, we observe that documents and similar objects feature the most part annotations per object. For example, bank statements, business cards, doctor's prescriptions, and similar objects feature the most part annotations per object. To ensure semantic labeling precision and increase granularity, we separated images featuring only the box versus the object and vice versa. In addition, we had comparable frame sampling across object labels, with every object label featuring between 650 and 800 human annotations per object. 

We observe a similar trend in cumulative part annotations per object as in object annotations per object illustrated in \textbf{Figure~\ref{fig:part-annots-per-obj}}. We observe that condom boxes and pregnancy test boxes have the lowest number of part annotations per object because these labels also have the lowest number of object annotations. We also note that condom packets, medical records, and pregnancy tests have the highest occurrences of single-part annotations. This is because the condom packet and pregnancy test only have two parts, where one part (branding) is predominantly more visible than the other (sensitive text) in nearly all viewing scenarios. The medical record has four parts; however, the patient data part label features the highest visibility across diverse viewing scenarios, as it occupies the most significant amount of area relative to the object compared to its other parts (diagnoses, prognoses, and physician info). The most common frequency of part annotations across all objects is between 1 and 4 part annotations per object. Lastly, we found that Bank statements, MIR reports, and Transcripts are the only objects featuring 6 potential parts; however, only bank statements and MIR reports have instances where all six parts were annotated, and transcripts do not. Tattoo sleeves and local newspapers most often show no part annotations, which we attribute to a lack of parts for tattoo sleeves and no visibility of the private content for local newspapers. 


\vspace{-0.5em}\paragraph{Segmentation Properties.} 
\begin{figure*}
    \centering
    \includegraphics[width=1.0\textwidth]{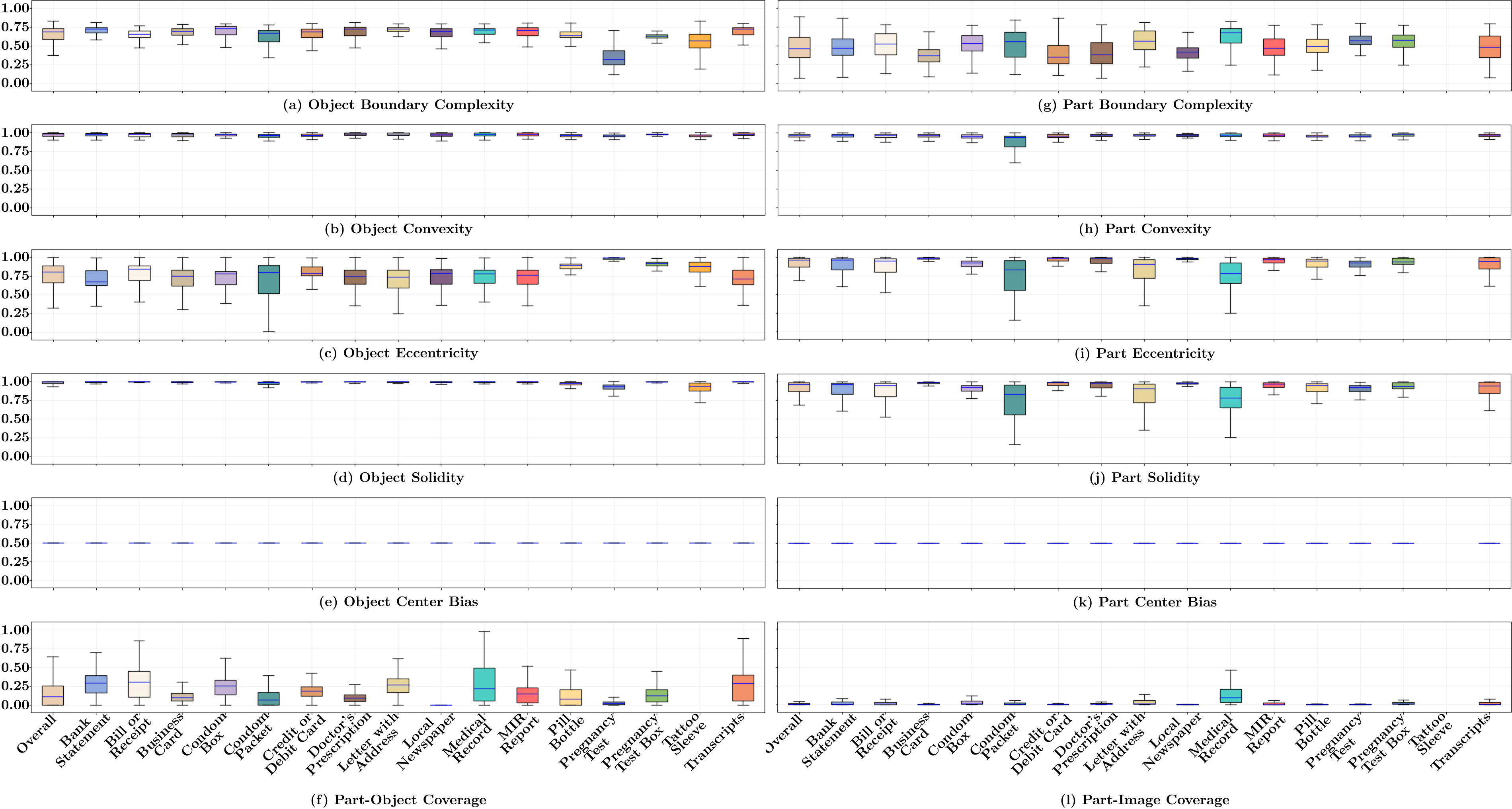}
    \vspace{-1.5em}
    \caption{Boxplots showing the distribution of boundary complexity, convexity, and eccentricity at the object and part level. Part-object coverage and part-image coverage are also shown. The blue lines represent medians, bottoms and tops of each box represent the 25th and 75th percentile values respectively, and whiskers represent the most extreme data points not considered outliers.}
     \label{fig:spatial_boxplots}
\end{figure*}

Statistics characterizing typical appearances of BIV-Priv-HIT's objects and parts are shown in \textbf{Figure~\ref{fig:spatial_boxplots}}. In BIV-Priv-HIT, we observe that most objects feature boundary complexities between 0.65 and 0.75 (\textbf{Figure \ref{fig:spatial_boxplots}a}), while most parts feature boundary complexities between 0.35 and 0.60 (\textbf{Figure \ref{fig:spatial_boxplots}e}). The pregnancy test object features the most jagged and diverse boundary complexity, with 75\% of its boundary complexities ranging from 0.25 to 0.42. We attribute this finding to pregnancy tests being the most geometrically complex objects out of all the object categories in the dataset. Moreover, the pregnancy test is the only object in the dataset that is not a square or rectangle and continuously presents complex boundaries regardless of the viewing angle. At the part level, the parts of the Business Card object feature the most jagged boundary complexities, with 75\% of values ranging from 0.29 to 0.44. We attribute this finding to the inherent jagged edges caused by the occurrence of `headings' and `information.' For example, business cards typically feature a heading such as `Job Title' or `Email' followed by the information, which is the actual job title or email address. In many cases, the information is longer than the heading, so when annotating the part where we directed annotators to include the heading, the information naturally lends itself to creating multiple jagged edges due to including the heading and information in a single part annotation. 

Regarding solidity (\textbf{Figure \ref{fig:spatial_boxplots}d}), nearly all objects and their respective parts are solid or `filled' (solidity values closer to 1), illustrating that nearly all objects and their parts are their own convex hulls, and exhibit minimal indentations in their perimeters. At the object level, we observe that nearly all objects feature solidity values ranging from 0.96 to 1.0, meaning that nearly all of the object's pixels also fall within its convex hull. The two notable exceptions to this observation are the pregnancy test. We attribute this finding to the pregnancy test being the most geometrically complex among the objects in the dataset, with the object's structure featuring several concavities. Similarly, the tattoo sleeve follows the shape of the arm from the elbow down to the wrist, lending itself to an inherently indented perimeter shape. We see slight variations in other objects but attribute these variations to viewing angles, occlusions, and other artifacts that can potentially alter the object's relative convexity, for example, viewing a document nearly straight on as opposed to from the top-down.

We see a similar phenomenon at the part level (\textbf{Figure \ref{fig:spatial_boxplots}j}); however, the objects with the more diverse solidity at the part level are the condom packet and the pill bottle. In the case of all the objects, we observe a similar trend to the object level: nearly all parts are `solid' with solidity values ranging from 0.95 to 1.0. Regarding the condom packet and the pill bottle, the exceptions to this trend, we attribute the increased convexity to the fact that these two objects feature a significant presence of text. In the case of the condom packet, the sensitive text is also placed among the branding, causing annotators to create more concavities in their annotations to segment sensitive text accurately. We see a similar trend in the pill bottle object due to parts such as addresses, personal information, and prescription data, all of which are shapes that require more significant concavities in their segmentation to accurately demarcate from the other parts. 

Regarding center bias (\textbf{Figure \ref{fig:spatial_boxplots}e}), values close to 0.5 indicate a balanced distribution of objects within the frame, suggesting that objects are neither heavily centralized nor significantly off-center. The dataset’s median center bias value is approximately 0.4997, and all objects feature a narrow center bias range between 0.49 and 0.5, indicating a precise and slight central tendency at the object level. At the part level (\textbf{Figure \ref{fig:spatial_boxplots}k}), we see the same median value of 0.4997, albeit the spreads and whiskers are slightly wider than the object level, indicating more variability in the positioning of parts within objects.

Regarding convexity (\textbf{Figure \ref{fig:spatial_boxplots}b}), we see similar trends to solidity at the object level, with convexity values ranging from 0.94 to 0.99. This finding suggests that the shapes are relatively smooth at the object level and lack significant indentations or concavities. We see almost an identical trend at the part level (\textbf{Figure \ref{fig:spatial_boxplots}h}), with convexity values typically ranging from 0.94 to 0.97. The only exception to this finding is the condom packet, which consists of two parts: sensitive text and branding. We find both of these parts to present many concavities due to the shapes required to segment sensitive text and branding accurately. For example, two of the condom packet brands found in the dataset are KY and Trojan; when segmenting the branding for these two condom packets, the KY logo and the Trojan helmet brand are shapes with many concavities and jagged edges. As a result, we observed that the parts of the condom packet had the broadest range of convexity values, ranging from 0.8 to 0.95.

Concerning eccentricity, at the object level (\textbf{Figure \ref{fig:spatial_boxplots}c}), we observe values exhibiting medians close to 0.8, a significant finding that indicates a generally high elongation in objects across categories. The interquartile range spans from approximately 0.62 to 0.98, further emphasizing the high median values. Moreover, we see a trend of whiskers extending from around 0.35 to 1.0, highlighting some eccentricity variation but maintaining a tendency towards higher values (more elongated). We see the condom packet's whiskers extend from 0.0 to 1.0, a finding we attribute primarily to the viewing angle because condom packets are only square-like when viewed top-down. In contrast, they can appear more elongated in nearly any other viewing scenario. We also observe high median values and tight spreads in the pill bottle, tattoo sleeve, and pregnancy test (median values $\geq$ 0.9), all of which are the most elongated objects in the dataset. 

At the part level (\textbf{Figure \ref{fig:spatial_boxplots}i}), we see similar trends, albeit with higher median values (0.75 to 0.98) and less variance compared to the object level (whiskers primarily between 0.5 and 1.0). Again, we observe the condom packet elicits the most diverse eccentricity values, mainly due to the placement of sensitive text and the unique shape of their graphical brandings. We see a similar phenomenon in the letter with the address and medical record objects, which we attribute to the presence of text as these two objects consist of the most textually dense parts compared to other objects in the dataset. Overall, the eccentricity values at the part level generally show less variation than the object level, as the object's parts tend to have more defined and consistent shapes within their parent objects. We also provide solidity and center bias statistics at the object and part levels, detailed further in the supplementary materials.  

Concerning part coverage, the relative area occupied by the region of interest, at the image level (\textbf{Figure \ref{fig:spatial_boxplots}l}) for nearly every object category, parts occupy less than 20\% of the image with a majority of parts occupying less than 5\% of the image. Again, we attribute the more significant inter-quartile range in the medical record object to the patient data, a part within medical records that can easily and often occupy more than half of the object.

At the object level (\textbf{Figure \ref{fig:spatial_boxplots}f}), we observe a similar phenomenon in that objects such as the bank statement, bill or receipt, medical record, and transcripts feature the largest interquartile ranges, a finding that we attribute to the relative sizes of the composite parts within these objects. For example, the transaction data part of a bank statement can and often does occupy most of the object compared to the account holder's name and address. Similarly, the grades part within the transcripts takes up most of the object's space instead of the student's name. In contrast, when examining the pregnancy test, we see a narrow interquartile range and a tight variance because the parts of the pregnancy test, such as the result and sensitive text, occupy very little space on the object itself. 

We also report findings for adopting size thresholds introduced for the MSCOCO dataset \cite{lin2014microsoft}, where 322 and 962 are thresholds determining whether an object is small, medium, or large.  We find that in BIV-Priv-HIT's object annotations, 0.1\% (13) of objects qualify as small, 2.9\% (298) as medium, and 95\% (9,854) as large. For part-level annotations, we find 6\% (1,323) qualify as small, 41\% (8,989) as medium, and 53\% (11,725) as large.  

\section{Model Benchmarking}

\begin{figure*}[t!] \centering
\includegraphics[width=\textwidth]{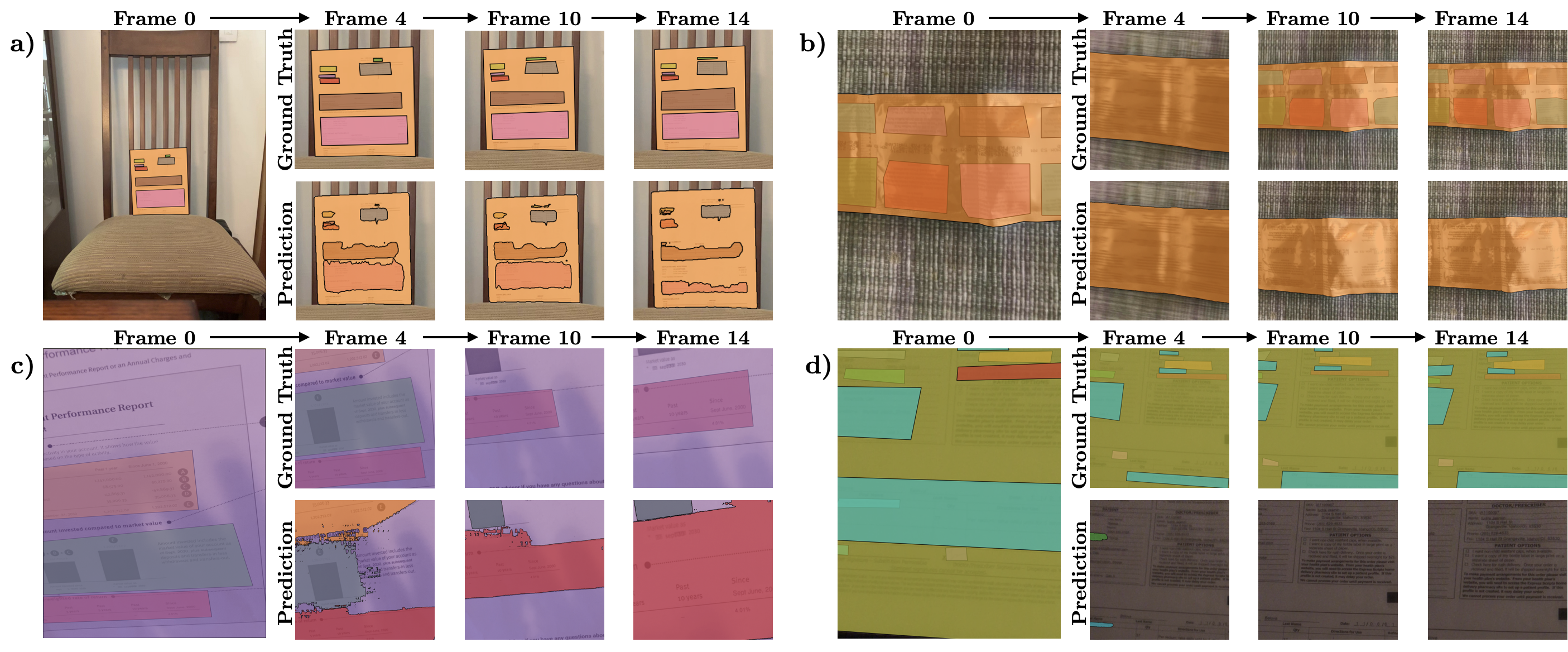}
    \vspace{-1em}
    \caption{Examples of SAM-2's performance on frames collected from four video clips in our dataset.  Shown is a full video frame with the ground truth mask (top) followed by cropped views of the ground truths and model predictions at subsequent frames in the video in order to make it easier to observe the model's performance on the region of interest. }\label{fig:qualitative-results-supps}
\end{figure*}

Despite the improved performance that comes from fine-tuning, our dataset still remains challenging for current state of the art models. \textbf{Figure~\ref{fig:qualitative-results-supps}a} shows that even a static object has so much variation in the predicted masks across frames. In \textbf{Figure~\ref{fig:qualitative-results-supps}b}, the model was unable to track all parts of the wrapper over time due to a shaky recording.  \textbf{Figure~\ref{fig:qualitative-results-supps}d} not only has more than nine parts, but the object is under low lighting, which appears together to be challenging for the model, even at the object level.

\end{document}